\documentclass[letterpaper]{article}

\usepackage{ijcai15}
\usepackage{times}

\usepackage{times}
\usepackage{helvet}
\usepackage{graphicx}
\usepackage{epstopdf}
\usepackage{todonotes}
\usepackage{danudefs}

\usepackage{algorithmic}
\usepackage{algorithm}

\usepackage{numprint}
\npdecimalsign{.}
\nprounddigits{2}

\usepackage{color}
\usepackage{colortbl}
\definecolor{LightCyan}{rgb}{0.88,1,1}

\newcommand{\analogy}{a\!:\!b,c\!:\!d}

\title{Embedding Semantic Relations into Word Representations}

\author{Danushka Bollegala \ \ Takanori Maehara   \ \ Ken-ichi Kawarabayashi\\
\ \ \ \ \  \ University of Liverpool \ \ Shizuoka University  \ \ National Institute of Informatics \\
JST, ERATO, Kawarabayashi Large Graph Project. 
}

\begin{document}

\maketitle

\begin{abstract}
Learning representations for semantic relations is important for various tasks
such as analogy detection, relational search, and relation classification.
Although there have been several proposals for learning representations for individual words,
learning word representations that explicitly capture the semantic relations between words remains under developed.
We propose an unsupervised method for learning vector representations for words
such that the learnt representations are sensitive to the semantic relations that exist between two words.
First, we extract lexical patterns from the co-occurrence contexts of two words
in a corpus to represent the semantic relations that exist between those two words.
Second, we represent a lexical pattern as the weighted sum of the representations of the words that co-occur with
that lexical pattern. Third, we train a binary classifier to detect relationally similar vs.
non-similar lexical pattern pairs.
The proposed method is unsupervised in the sense that the lexical pattern pairs we use as train data
are automatically sampled from a corpus, without requiring any manual intervention.
Our proposed method statistically significantly outperforms
the current state-of-the-art word representations on three benchmark datasets for
proportional analogy detection, demonstrating its ability to accurately capture the semantic relations among words.
\end{abstract}

\section{Introduction}
\label{sec:intro}

Representing the semantics of words and relations are fundamental tasks in Knowledge Representation (KR).
Numerous methods for learning distributed word representations have been proposed 
in the NLP community~\cite{Turian:ACL:2010,Collobert:2011,Mikolov:2013,Milkov:2013,Pennington:EMNLP:2014}.
Distributed word representations have shown to improve performance in a wide-range of tasks such as,
machine translation~\cite{cho-EtAl:2014:EMNLP2014}, semantic similarity measurement~\cite{Mikolov:NAACL:2013,Pennington:EMNLP:2014},
and word sense disambiguation~\cite{Huang:ACL:2012}. 

Despite the impressive performance of representation learning methods for individual words,
existing methods use only co-occurrences between words, ignoring the rich semantic relational structure.
The context in which two words co-occur provides useful insights into the semantic relations that exist between those two words.
For example, the sentence \emph{\textbf{ostrich} is a large \textbf{bird}} not only provides the information that
\textit{ostrich} and \textit{bird} are co-occurring, but also describes how they are related via the lexical pattern
\textbf{X} \textit{is a large} \textbf{Y}, where slot variables \textbf{X} and \textbf{Y} correspond to the two words between which
the relation holds. If we can somehow embed the information about the semantic relations $R$ that are associated with a particular
word $w$ into the representation of $w$, then we can construct richer semantic representation than the pure
co-occurrence-based word representations.
Although the word representations learnt by co-occurrence prediction methods~\cite{Mikolov:NAACL:2013,Pennington:EMNLP:2014}
have implicitly captured a certain degree of relational structure, it remains unknown how to explicitly embed the information about
semantic relations into word representations.

We propose a method for learning word representations that explicitly encode the information about
the semantic relations that exist between words.
Given a large corpus, we extract lexical patterns that correspond to numerous semantic relations
that exist between word-pairs $(x_i, x_j)$. 
Next, we represent each word $x_i$ in the vocabulary by a $d$-dimensional vector $\vec{x}_i \in \R^d$.
Word representations can be initialized either randomly or by using pre-trained word representations.
Next, we represent a pattern $p$ by the weighted average of the vector differences $(\vec{x}_i - \vec{x}_j)$ corresponding to
word-pairs $(x_i, x_j)$ that co-occur with $p$ in the corpus. This enables us to represent a pattern $p$
by a $d$-dimensional vector $\vec{p} \in \R^d$ in the same embedding space as the words.
Using vector difference between word representations to represent semantic relations is motivated by the
observations in prior work on word representation learning~\cite{Mikolov:NAACL:2013,Pennington:EMNLP:2014}
where, for example,  the difference of vectors representing \textit{king} and \textit{queen}
has shown to be similar to the difference of vectors  representing \textit{man} and \textit{woman}.

We model the problem of embedding semantic relations into word representations 
as an analogy prediction task where, given two lexical patterns, we train a binary classifier that predicts 
whether they are relationally similar.
Our proposed method is unsupervised in the sense that 
both positive and negative training instances that we use for training are automatically
selected from a corpus, without requiring any manual intervention.
Specifically, pairs of lexical patterns that co-occur with the same set of word-pairs are selected as positive training
instances, whereas negative training instances are randomly sampled from pairs of patterns with low relational similarities.
Our proposed method alternates between two steps (Algorithm~\ref{algo:relrep}). In the first step, we construct pattern representations from 
current word representations. In the second step, we predict whether a given pair of patterns is relationally similar
using the computed representations of patterns in the previous step.
We update the word representations such that the prediction loss is minimized.

Direct evaluation of word representations is difficult because there is no agreed gold standard for semantic representation of words. 
Following prior work on representation learning, we evaluate the proposed method 
using the learnt word representations in an analogy detection task. 
For example, denoting the word representation for a word $w$ by $\vec{v}(w)$, the vector
$\vec{v}({\rm king}) - \vec{v}({\rm man}) + \vec{v}({\rm woman})$ is required to be similar to $\vec{v}({\rm queen})$,
than all the other words in the vocabulary. Similarity between two vectors is computed by the cosine of the angle between the
corresponding vectors. 
The accuracy obtained in the analogy detection task with a particular word representation method
is considered as a measure of its accuracy.
In our evaluations, we use three previously proposed benchmark datasets for 
word analogy detection: SAT analogy dataset~\cite{Turney_LRA}, 
Google analogy dataset~\cite{Mikolov:NIPS:2013}, 
and SemEval analogy dataset~\cite{SemEavl2012:Task2}.
The word representations produced by our
proposed method statistically significantly outperform the current state-of-the-art word representation learning methods
on all three benchmark datasets in an analogy detection task, demonstrating the accuracy of the proposed
method for embedding semantic relations in word representations.

\section{Related Work}

Representing words using vectors (or tensors in general) is an essential task in text processing. 
For example, in distributional semantics~\cite{Baroni:DM}, a word $x$ is represented by a vector that contains other words that
co-occur with $x$ in a corpus. Numerous methods for selecting co-occurrence contexts
(e.g. proximity-based windows, dependency relations), and word association measures
(e.g. pointwise mutual information (PMI), log-likelihood ratio (LLR), local mutual information (LLR)) have been proposed~\cite{Turney:JAIR:2010}. Despite the successful applications of co-occurrence counting-based distributional word representations,
their high dimensionality and sparsity is often problematic when applied in NLP tasks.
Consequently, further post-processing such as dimensionality reduction,
and feature selection is often required when using distributional word representations.

On the other hand, distributed word representation learning methods model words as $d$-dimensional real vectors
and learn those vector representations by applying them to solve an auxiliary task such as language modeling.
The dimensionality $d$ is fixed for all the words in the vocabulary and,
unlike distributional word representations, is much smaller (e.g. $d \in [10, 1000]$ in practice) compared to the vocabulary size.
A pioneering work on word representation learning is the neural network language model (NNLMs)~\cite{Bengio:JMLR:2003},
where word representations are learnt such that we can accurately predict the next word in 
a sentence using the word representations for the previous words. 
Using backpropagation, word vectors are updated such that the prediction error is minimized.

Although NNLMs learn word representations as a by-product, the main focus on language modeling is
to predict the next word in a sentence given the previous words, and not on learning word representations that capture word semantics.
Moreover, training multi-layer neural networks with large text corpora is often time consuming.
To overcome those limitations, methods that specifically focus on learning word
representations that capture word semantics using large text corpora have been proposed.
Instead of using only the previous words in a sentence as in language modeling,
these methods use \emph{all} the words in a contextual window for the prediction task~\cite{Collobert:2011}.
Methods that use one or no hidden layers are proposed to improve the scalability of the learning algorithms.
For example, the skip-gram model~\cite{Mikolov:NIPS:2013} predicts the words $c$ that appear in the local context of a word $x$, whereas
the continuous bag-of-words model (CBOW) predicts a word $x$ conditioned on all the words $c$ that appear in
$x$'s local context~\cite{Milkov:2013}.
However, methods that use global co-occurrences in the entire corpus to learn word
representations have shown to outperform methods that use only local co-occurrences~\cite{Huang:ACL:2012,Pennington:EMNLP:2014}.
Word representations learnt using above-mentioned representation learning methods have shown superior performance over word representations
constructed using the traditional counting-based methods~\cite{baroni-dinu-kruszewski:2014:P14-1}.

Word representations can be further classified depending on whether they are task-specific or task-independent.
For example, methods for learning word representations for specific tasks such as sentiment classification~\cite{socher-EtAl:2011:EMNLP}, and
semantic composition~\cite{hashimoto-EtAl:2014:EMNLP2014} have been proposed.
These methods use label data for the target task to train supervised models, and learn word representations
that optimize the performance on this target task. 
Whether the meaning of a word is task-specific or task-independent remains an interesting open question.
Our proposal can be seen as a third alternative in the sense that we use task-independent pre-trained
word representations as the input, and embed the knowledge related to the semantic relations into the word representations.
However, unlike the existing task-specific word representation learning methods, we do not require manually
labeled data for the target task (i.e. analogy detection).

\section{Learning Word Representations}

The local context in which two words co-occur provides useful information regarding the semantic relations that
exist between those two words. For example, from the sentence \emph{\textbf{Ostrich} is a large \textbf{bird} that 
primarily lives in Africa}, we can infer that the semantic relation \textsf{IS-A-LARGE} exists between \textit{ostrich}
and \textit{bird}. Prior work on relational similarity measurement have successfully used such lexical patterns as features
to represent the semantic relations that exist between two words~\cite{Duc:WI:2010,Duc:AAAI:2011}.
According to the \emph{relational duality hypothesis}~\cite{Bollegala:WWW:2010},
a semantic relation $R$ can be expressed either \emph{extensionally} by enumerating word-pairs for which $R$ holds, or
\emph{intensionally} by stating lexico-syntactic patterns that define the properties of $R$.

Following these prior work, we extract lexical patterns from the co-occurring contexts of two words to represent the semantic
relations between those two words. Specifically, we extract unigrams and bigrams of tokens as patterns from the \emph{midfix}
(i.e. the sequence of tokens that appear in between the given two words in a context)~\cite{Bollegala:JSAI:2007,Bollegala:NAACL:2007}.
Although we use lexical patterns as features for representing semantic relations in this work, our proposed method
is not limited to lexical patterns, and can be used in principle with any type of features that represent relations.
The strength of association between a word pair $(u,v)$ and
a pattern $p$ is measured using the positive pointwise mutual information (PPMI), $f(p, u, v)$, which is defined as follows,
\begin{equation}
\label{eq:PPMI}
\small
f(p,u,v) = \max(0, \log \left( \frac{g(p,u,v) g(*,*,*)}{g(p,*,*) g(*,u,v)} \right)) .
\end{equation}
Here, $g(p,u,v)$ denotes the number of co-occurrences between $p$ and $(u,v)$, and $*$ denotes the summation
taken over all words (or patterns) corresponding to the slot variable.
We represent a pattern $p$ by the set $\cR(p)$ of word-pairs $(u,v)$ for which  $f(p,u,v) > 0$.
Formally, we define $\cR(p)$ and its norm $|\cR(p)|$ as follows,
\begin{eqnarray}
\label{eq:R}
\small
\cR(p) = \{(u,v) | f(p,u,v) > 0 \} \\
\label{eq:R-norm}
\small
|\cR(p)| = \sum_{(u,v) \in \cR(p)} f(p,u,v) 
\end{eqnarray}
We represent a word $\vec{x}$ using a vector $\vec{x} \in \R^d$.
The dimensionality of the representation, $d$, is a hyperparameter of the proposed method. 
Prior work on word representation learning have observed that
the difference between the vectors that represent two words closely approximates the semantic relations that exist between
those two words. For example, the vector $\vec{v}(\textrm{king}) - \vec{v}(\textrm{queen})$ has shown to be similar to the
vector $\vec{v}(\textrm{man}) - \vec{v}(\textrm{woman})$. We use this property to represent a pattern $p$
by a vector $\vec{p} \in \R^d$ as the weighted sum of differences between the two words in all word-pairs $(u,v)$ that co-occur
with $p$ as follows,
\begin{equation}
\small
\label{eq:p}
\vec{p} = \frac{1}{|\cR(p)|} \sum_{(u,v) \in \cR(p)} f(p,u,v)(\vec{u} - \vec{v}) .
\end{equation}

\begin{figure}[t]
\centering
\includegraphics[width=80mm]{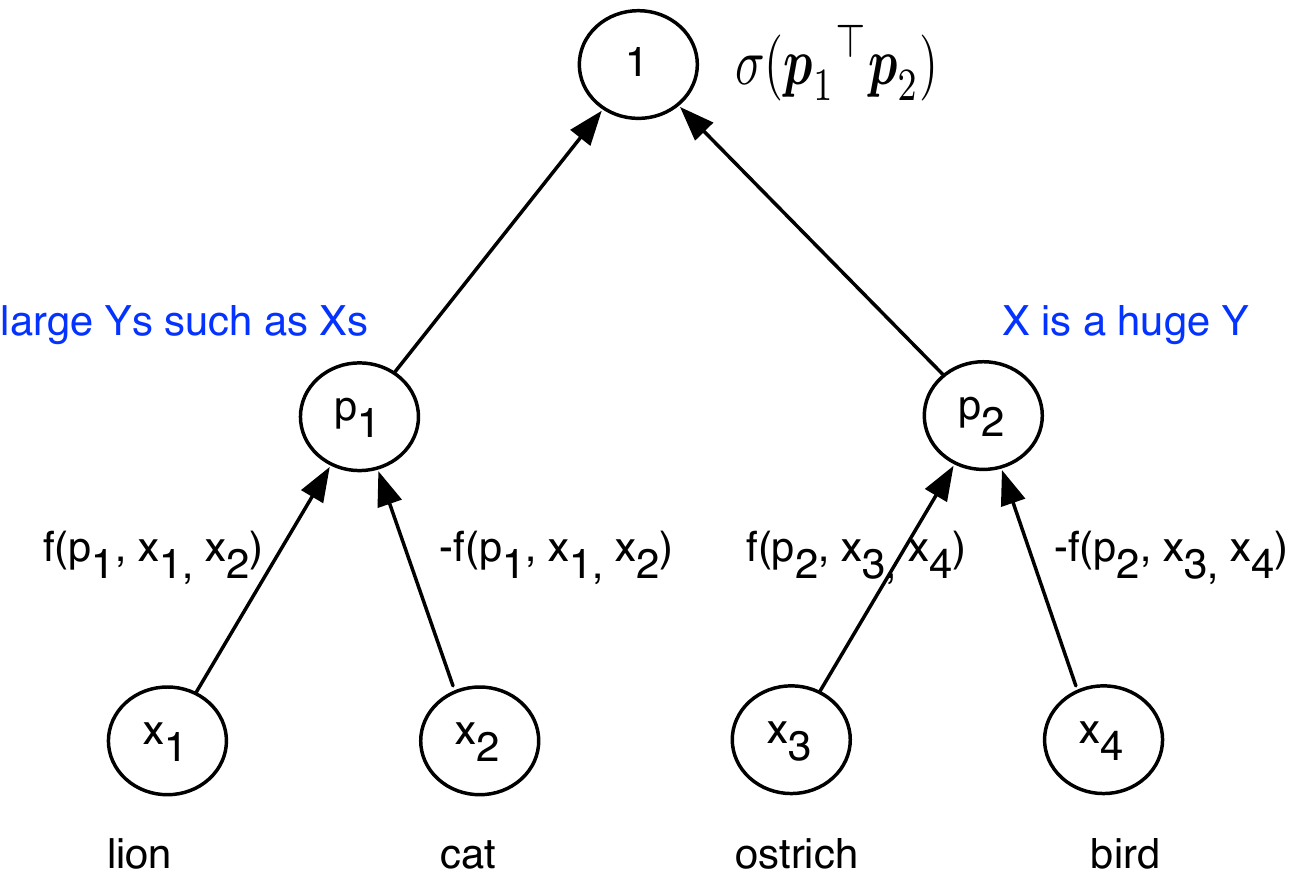}
\caption{Computing the similarity between two patterns.}
\label{fig:outline}
\end{figure}

For example, consider Fig.~\ref{fig:outline}, where the two word-pairs 
(\textit{lion}, \textit{cat}), and (\textit{ostrich}, \textit{bird}) co-occur respectively with the two lexical patterns,
$p_1 = \textrm{\textit{large \textbf{Y}s such as \textbf{X}s}}$, 
and $p_2 = \textrm{\textit{\textbf{X} is a huge \textbf{Y}}}$.
Assuming that there are no other co-occurrences between word-pairs
and patterns in the corpus, the representations of the patterns $p_1$ and $p_2$ are given respectively by
$\vec{p}_1 = \vec{x}_1 - \vec{x}_2$, and $\vec{p}_2 = \vec{x}_3 - \vec{x}_4$.
We measure the relational similarity between $(x_1, x_2)$ and $(x_3, x_4)$ using the inner-product ${\vec{p}_1}\T\vec{p}_2$.

We model the problem of learning word representations as a binary classification task, where we
learn representations for words such that they can be used to accurately predict whether
a given pair of patterns are relationally similar. 
In our previous example, we would learn representations for the four words
\emph{lion}, \emph{cat}, \emph{ostrich}, and \emph{bird} such that the similarity between the two patterns
\textit{large \textbf{Y}s such as \textbf{X}s}, and \textit{\textbf{X} is a huge \textbf{Y}} is maximized.
Later in Section \ref{sec:train:data}, we propose an unsupervised method for selecting relationally similar (positive) and dissimilar 
(negative) pairs of patterns as training instances to train a binary classifier.

Let us denote the target label for two patterns $p_1$, $p_2$ by $t(p_1, p_2) \in \{1, 0\}$, where the value $1$
indicates that $p_1$ and $p_2$ are relationally similar, and $0$ otherwise. 
We compute the prediction loss for a pair of patterns $(p_1,p_2)$ as the squared loss between the target
and the predicted labels as follows,
\begin{equation}
\label{eq:loss}
L(t(p_1, p_2), p_1, p_2) = \frac{1}{2}{(t(p_1, p_2) - \sigma({\vec{p}_1}\T\vec{p}_2))}^2.
\end{equation}
Different non-linear functions can be used as the prediction function $\sigma(\cdot)$
such as the logistic-sigmoid, hyperbolic tangent, or rectified linear units.
In our preliminary experiments we found hyperbolic tangent, $\tanh$, given by
\begin{equation}
\label{eq:tanh}
\small
\sigma(\theta) = \tanh(\theta) = \frac{\exp(\theta) - \exp(-\theta)}{\exp(\theta) + \exp(-\theta)}
\end{equation}
to work particularly well among those different non-linearities.

To derive the update rule for word representations, let us consider the derivative of the loss w.r.t. the word representation
$\vec{x}$ of a word $x$,
\begin{equation}
\small
\label{eq:L/x}
\frac{\partial L}{\partial \vec{x}} = \frac{\partial L}{\partial \vec{p}_1}\frac{\partial \vec{p}_1}{\partial \vec{x}} + 
							   \frac{\partial L}{\partial\vec{p}_2}\frac{\partial \vec{p}_2}{\partial \vec{x}} ,
\end{equation}
where the partial derivative of the loss w.r.t. pattern representations are given by,
\begin{eqnarray}
\small
\label{eq:L/p1}
\frac{\partial L}{\partial \vec{p}_1} = \sigma'(\vec{p}_1\T\vec{p}_2) (\sigma(\vec{p}_1\T\vec{p}_2) - t(p_1,p_2)) \vec{p}_2, \\
\label{eq:L/p2}
\small
\frac{\partial L}{\partial \vec{p}_2} = \sigma'(\vec{p}_1\T\vec{p}_2) (\sigma(\vec{p}_1\T\vec{p}_2) - t(p_1,p_2)) \vec{p}_1 .
\end{eqnarray}
Here, $\sigma'$ denotes the first derivative of $\tanh$, which is given by $1 - {\sigma(\theta)}^2$.
To simplify the notation we drop the arguments of the loss function.

From Eq.~\ref{eq:p} we get,
\begin{eqnarray}
\label{eq:p1/x}
\small
\frac{\partial \vec{p_1}}{\partial \vec{x}} = \frac{1}{|\cR(p_1)|} \left( h(p_1, u = x, v) - h(p_1, u, v = x) \right), \\
\label{eq:p2/x}
\small
\frac{\partial \vec{p_2}}{\partial \vec{x}} = \frac{1}{|\cR(p_2)|} \left( h(p_2, u = x, v) - h(p_2, u, v = x) \right),
\end{eqnarray}
where,
\begin{equation*}
\small
h(p, u = x, v) = \sum_{(x,v) \in \{(u,v) | (u,v) \in \cR(p), u = x\}} f(p, x, v), 
\end{equation*}
and
\begin{equation*}
h(p, u, v = x) = \sum_{(u,x) \in \{(u,v) | (u,v) \in \cR(p), v = x\}} f(p, u, x) .
\end{equation*}

Substituting the partial derivatives given by Eqs.~\ref{eq:L/p1}-\ref{eq:p2/x} in Eq.~\ref{eq:L/x} we get,
\begin{eqnarray}
\nonumber
\small
\label{eq:grad}
\frac{\partial L}{\partial \vec{x}} = \lambda(p_1, p_2)  [ H(p_1,x)\sum_{(u,v) \in \cR(p_2)} f(p_2,u,v) (\vec{u}-\vec{v}) \\ + H(p_2, x) \sum_{(u,v) \in \cR(p_1)} f(p_1,u,v) (\vec{u}-\vec{v}) ],
\end{eqnarray}
where $\lambda(p_1,p_2)$ is defined as
\begin{equation}
\small
\lambda(p_1, p_2) = \frac{\sigma'(\vec{p}_1\T\vec{p}_2) (t(p_1,p_2) - \sigma(\vec{p}_1\T\vec{p}_2))}{|\cR(p_1)||\cR(p_2)|},
\end{equation}
and $H(p,x)$ is defined as
\begin{equation}
\small
H(p,x) = h(p, u=x, v) - h(p, u, v=x) .
\end{equation}

\begin{algorithm}[t]       
\caption{Learning word representations.}        
\label{algo:relrep}                         
\begin{algorithmic}[1]         
\REQUIRE Training pattern-pairs $\{(p^{(i)}_1, p^{(i)}_2, t(p^{(i)}_1, p^{(i)}_2)\}_{i=1}^{N}$, dimensionality $d$ of the word representations,
 and the maximum number of iterations $T$.
\ENSURE Representation $\vec{x}_j \in \R^d$, of a word $x_j$ for $j=1, \ldots, M$, where $M$ is the vocabulary size.
\medskip
\STATE Initialize word vectors $\{\vec{x}_j\}_{j=1}^M$. \label{line:init}
\FOR{ $t = 1$ \TO $T$ }
	\FOR{$k = 1$ \TO $K$}
		\STATE $\vec{p}_k = \frac{1}{|\cR(p_k)|} \sum_{(u,v) \in \cR(p_k)} f(p_k,u,v) (\vec{u} - \vec{v})$ \label{line:pats}
	\ENDFOR
	\FOR{$i=1$ \TO $N$}
		\FOR{$j=1$ \TO $M$}
			\STATE $\vec{x}_j = \vec{x}_j - \alpha^{(t)}_j \frac{\partial L}{\partial \vec{x}_j}$ \label{line:update}
		\ENDFOR
	\ENDFOR 
\ENDFOR 
\RETURN $\{\vec{x}_j\}_{j=1}^M$. 
\end{algorithmic}
\end{algorithm}

We use stochastic gradient decent (SGD) with learning rate adapted by AdaGrad~\cite{Duchi:JMLR:2011}
to update the word representations. The pseudo code for the proposed method is shown in
Algorithm~\ref{algo:relrep}. Given a set of $N$ relationally similar and dissimilar pattern-pairs,
 $\{(p^{(i)}_1, p^{(i)}_2, t(p^{(i)}_1, p^{(i)}_2)\}_{i=1}^{N}$, Algorithm~\ref{algo:relrep} initializes each word $x_j$ in the vocabulary
 with a vector $\vec{x}_j \in \R^d$. The initialization can be conducted either using randomly sampled vectors from 
 a zero mean and unit variance Gaussian distribution, or by pre-trained word representations.
 In our preliminary experiments, we found that the word vectors learnt by
GloVe~\cite{Pennington:EMNLP:2014} to perform consistently well over random vectors when used 
as the initial word representations in the proposed method.
Because word vectors trained using existing word representations already demonstrate
a certain degree of relational structure with respect to proportional analogies, we believe that initializing using pre-trained
word vectors assists the subsequent optimization process. 

During each iteration, Algorithm~\ref{algo:relrep} alternates between two steps.
 First, in Lines 3-5, it computes pattern representations using Eq.~\ref{eq:p} from the current word representations
 for all the patterns ($K$ in total) in the training dataset. Second, in Lines 6-10, for each train pattern-pair
we compute the derivative of the loss according to Eq.~\ref{eq:grad}, and update the word representations.
These two steps are repeated for $T$ iterations, after which the final set of word representations are returned.

The computational complexity of Algorithm~\ref{algo:relrep} is $\O(TKd + TNMd)$,
where $d$ is the dimensionality of the word representations.
Naively iterating over $N$ training instances and $M$ words in the vocabulary 
can be prohibitively expensive for large training datasets and vocabularies.
However, in practice we can efficiently compute the updates using two tricks:  \emph{delayed updates}
and \emph{indexing}. Once we have computed the pattern representations for all $K$ patterns
in the first iteration, we can postpone the update of a representation for a pattern until that pattern
next appears in a training instance. This reduces the number of patterns that are updated in each iteration to a maximum of $2$ instead of $K$
for the iterations $t > 1$. Because of the sparseness in co-occurrences, only a handful (ca. $100$) of patterns co-occur with any given word-pair.
Therefore, by pre-compiling an index from a pattern to the words with which that pattern co-occurs, we can limit the
update of word representations in Line~\ref{line:update} to a much smaller number than $M$.
Moreover, the vector subtraction can be parallized across the dimensions.
Although the loss function defined by Eq.~\ref{eq:loss} is non-convex w.r.t. to word representations,
in practice, Algorithm~\ref{algo:relrep} converges after a few (less than $5$) iterations. 
In practice, it requires less than an hour to train from a 2 billion word corpus where we have
$N=100,000$, $T=10$, $K=10,000$ and $M=210,914$.

Lexical patterns contain sequences of multiple words. 
Therefore, exact occurrences of lexical patterns are rare compared to that of individual words even in large corpora.
Directly learning representations for lexical patterns using their co-occurrence statistics leads to data sparseness issues,
which becomes problematic when applying existing methods proposed for learning representations for single words
to learn representations for lexical patterns that consist of multiple words.
The proposal made in Eq.~\ref{eq:p} to compute representations for patterns circumvent this data sparseness issue by
indirectly modeling patterns through word representations.

\subsection{Selecting Similar/Dissimilar Pattern-Pairs}
\label{sec:train:data}

We use the ukWaC corpus\footnote{\url{http://wacky.sslmit.unibo.it}}
to extract relationally similar (positive) and dissimilar (negative) pairs of patterns $(p_i, p_j)$ 
to train the proposed method. The ukWaC is a 2 billion word corpus
constructed from the Web limiting the crawl to the \textbf{.uk} domain.
We select word-pairs that co-occur at least in $50$ sentences within a co-occurrence window of $5$ tokens.
Moreover, using a stop word list, we ignore word-pairs that purely consists of stop words.
We obtain $210,914$ word-pairs from this step. Next, we extract lexical patterns for those word-pairs
by replacing the first and second word in a word-pair respectively by slot variables
\textbf{X} and \textbf{Y} in a co-occurrence window of length $5$ tokens to extract numerous lexical patterns.
We select the top occurring $10,000$ lexical patterns (i.e. $K = 10,000$) for further processing.

We represent a pattern $p$ by a vector where the elements correspond to the PPMI values $f(p, u, v)$ between $p$
and all the word-pairs $(u,v)$ that co-occur with $p$. Next, we compute the cosine similarity between all pairwise
combinations of the $10,000$ patterns, and rank the pattern pairs in the descending order of their
cosine similarities. We select the top ranked $50,000$ pattern-pairs as positive training instances.
We select $50,000$ pattern-pairs from the bottom of the list which have non-zero similarity scores
as negative training instances. The reason for not selecting pattern-pairs with zero similarity scores
is that such patterns do not share any word-pairs in common, and are not informative as training data
for updating word representations. Thus, the total number of training instances we select is
$N = 50,000 + 50,000 = 100,000$.

\section{Evaluating Word Representations using Proportional Analogies}
\label{sec:eval}

To evaluate the ability of the proposed method to learn word representations that embed information related to
semantic relations, we apply it to detect proportional analogies.
For example, consider the proportional analogy,
\textit{man}:\textit{woman} :: \textit{king}:\textit{queen}. Given, the first three words, a word representation learning
method is required to find the fourth word from the vocabulary that maximizes the relational similarity between
the two word-pairs in the analogy. Three benchmark datasets have been popularly used in prior work for evaluating analogies: 
\textbf{Google} dataset~\cite{Mikolov:NIPS:2013} ($10,675$ syntactic analogies and $8869$ semantic analogies),
\textbf{SemEval} dataset~\cite{SemEavl2012:Task2} ($79$ questions), and
\textbf{SAT} dataset~\cite{Turney_CL} ($374$ questions).
For the Google dataset, the set of candidates for the fourth word consists of all the words in the vocabulary.
For the SemEval and SAT datasets, each question word-pair is assigned with a limited number of candidate word-pairs out of which
only one is correct. The accuracy of a word representation is evaluated by the percentage of the correctly answered
analogy questions out of all the questions in a dataset. We do not skip any questions in our evaluations.

Given a proportional analogy $a:b :: c:d$, we use the following measures proposed in prior work
for measuring the relational similarity between $(a,b)$ and $(c,d)$.
\begin{description}
\itemsep=0pt
\item[CosAdd]
proposed by Mikolov et al.~\shortcite{Mikolov:NAACL:2013} ranks candidates $c$ according to the formula
\begin{equation}
\label{eq:cosadd}
\small
{\rm CosAdd}(\analogy) = \cos(\vec{b}-\vec{a}+\vec{c}, \vec{d}),
\end{equation}
and selects the top-ranked candidate as the correct answer.

\item[CosMult:] CosAdd measure can be decomposed into the summation of three cosine similarities, where in practice one of the three terms
often dominates the sum. To overcome this bias in CosAdd, Levy and Goldberg~\shortcite{Levy:CoNLL:2014} proposed
the \textbf{CosMult} measure given by,
\begin{equation}
\label{eq:cosmult}
\small
{\rm CosMult}(\analogy) = \frac{\cos(\vec{b},\vec{d}) \cos(\vec{c},\vec{d})}{\cos(\vec{a},\vec{d}) + \epsilon} .
\end{equation}
We convert all cosine values $x \in [-1,1]$ to positive values using the transformation $(x + 1)/2$.
Here, $\epsilon$ is a small constant value to prevent denominator becoming zero, and is set to $10^{-5}$ in the experiments.

\item[PairDiff] measures the cosine similarity between the two vectors that correspond to the difference of the
word representations of the two words in each word-pair. It follows from our hypothesis that the semantic relation between
two words can be represented by the vector difference of their word representations.
PairDiff has been used by Mikolov et al.~\shortcite{Mikolov:NAACL:2013} for detecting semantic analogies
and is given by,
\begin{equation}
\label{eq:pairdiff}
\small
{\rm PairDiff}(\analogy) = \cos(\vec{b}-\vec{a}, \vec{d} - \vec{c}) .
\end{equation}
\end{description}

\section{Experiments and Results}
\label{sec:exp}

\begin{table}[t!]
\small
\centering
\caption{Word analogy results on benchmark datasets.}
\label{tbl:baselines}
\begin{tabular}{|l||p{6mm}|p{6mm}|p{6mm}|p{6mm}|p{10mm}|}
\hline 
\rowcolor{LightCyan}
Method 				& sem. 		& synt. 		& total 		& SAT 		& SemEval \\ \hline

ivLBL CosAdd		& 63.60		&	61.80	&	62.60	 	&	20.85	&	34.63	\\
ivLBL CosMult		& 65.20		&	63.00	&	64.00	&	19.78	&	33.42	\\	
ivLBL PairDiff			& 52.60		&	48.50	&	50.30	&	22.45	&	36.94 \\ \hline
	
skip-gram CosAdd		&	31.89	&	67.67	&	51.43	&	29.67	&	40.89 \\
skip-gram CosMult	&	33.98	&	69.62	&	53.45	&	28.87	&	38.54 \\
skip-gram PairDiff		&	7.20		&	19.73	&	14.05	&	35.29	&	43.99 \\ \hline
CBOW CosAdd		&	39.75	&	70.11	&	56.33	&	29.41	&	40.31 \\
CBOW CosMult		&	38.97	&	70.39	&	56.13	&	28.34	&	38.19 \\
CBOW PairDiff		&	5.76		&	13.43	&	9.95		&	33.16	&	42.89 \\ \hline
GloVe CosAdd			&	86.67 	& 	82.81 	& 	84.56 	& 	27.00 	& 	40.11 \\
GloVe CosMult		&	86.84	& 	84.80 	& 	85.72 	& 	25.66 	& 	37.56 \\
GloVe PairDiff			&	45.93	&	41.23	&	43.36	&	44.65	&	44.67 \\  \hline \hline

Prop CosAdd 			&	86.70		 & 85.35		 	& 85.97		 	& 29.41		 	& 41.86  \\
Prop CosMult 		&	\textbf{86.91} & \textbf{87.04} & \textbf{86.98} 	& 28.87 			& 39.67 \\
Prop PairDiff 			&  41.85			 & 42.86			 & 42.40			 & \textbf{45.99} 	& \textbf{44.88} \\ 
\hline \end{tabular}
\end{table}

In Table~\ref{tbl:baselines}, we compare the proposed method against previously proposed word representation learning methods:
\textbf{ivLBL}~\cite{Mnih:2013}, \textbf{skip-gram}~\cite{Mikolov:NIPS:2013}, \textbf{CBOW}~\cite{Milkov:2013}, 
and \textbf{GloVe}~\cite{Pennington:EMNLP:2014}.
All methods compared in Table~\ref{tbl:baselines} are trained on the same ukWaC corpus of 2B tokens to produce $300$ dimensional
word vectors. We use the publicly available implementations\footnote{\url{https://code.google.com/p/word2vec/}}\textsuperscript{,}\footnote{\url{http://nlp.stanford.edu/projects/glove/}}  by the original authors for training the word representations using the recommended parameter values.
Therefore, any differences in performances reported in Table~\ref{tbl:baselines} can be directly attributable to the
differences in the respective word representation learning methods.
In all of our experiments, the proposed method converged with less than $5$ iterations.

\begin{figure}[t]
\begin{center}
\includegraphics[height=60mm]{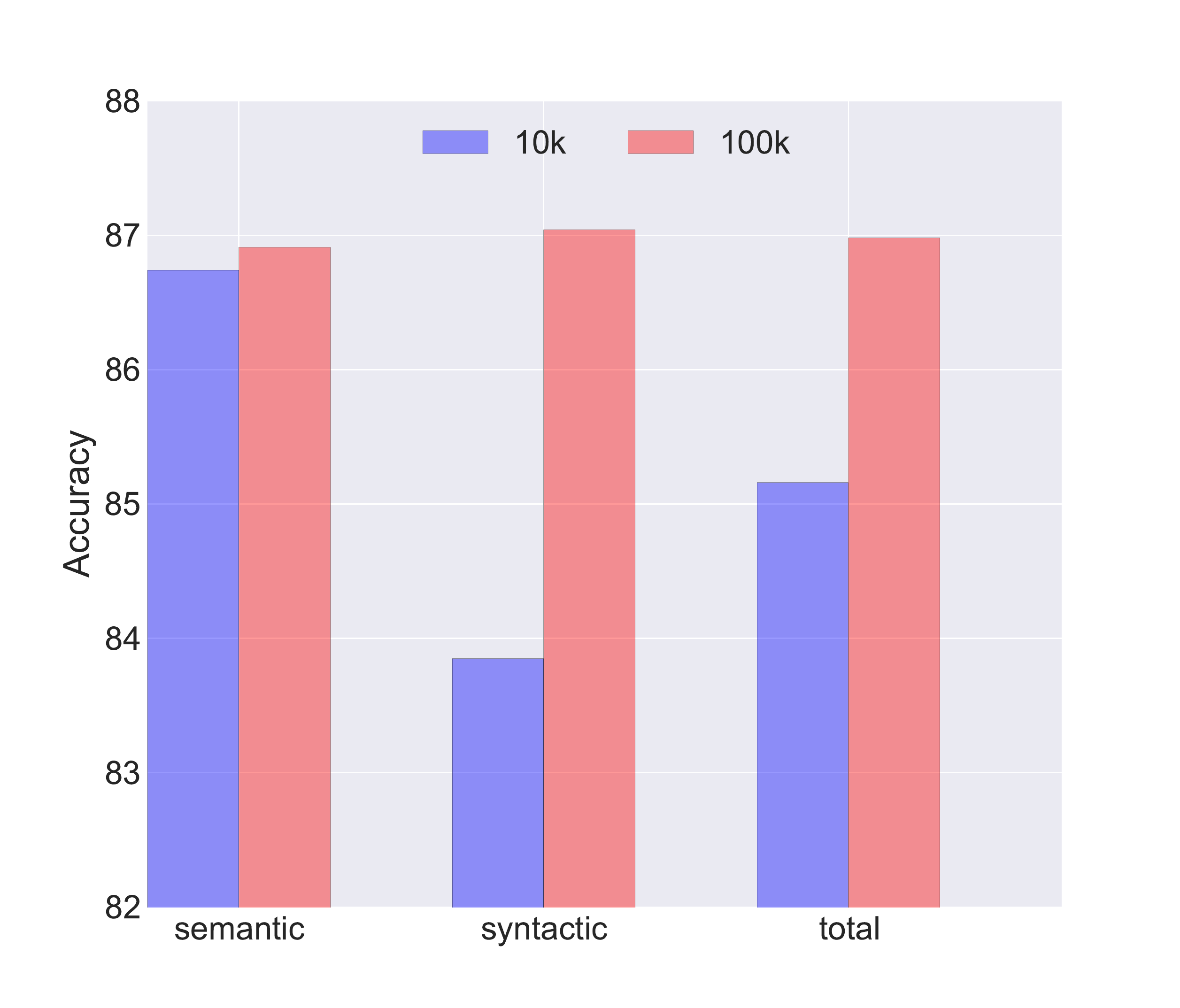}
\caption{Accuracy on Google dataset when the proposed method is trained using 10k and 100k instances.}
\label{fig:datasetsize}
\end{center}
\end{figure}

From Table~\ref{tbl:baselines} we see that the proposed method (denoted by \textbf{Prop}) achieves the best results
for the semantic (\textbf{sem}), syntactic (\textbf{synt}) and their union (\textbf{total}) analogy questions in the Google dataset using CosMult measure.
For analogy questions in SAT and SemEval datasets the best performance is reported by the proposed method using the PairDiff measure.
The PairDiff measure computes the cosine similarity between the two difference vectors 
$\vec{b}-\vec{a}$ and $\vec{d} - \vec{c}$, ignoring the spatial distances between the individual words as opposed to
CosAdd or CosMult. Recall that in the Google dataset we are required to find analogies from a large open vocabulary
whereas in SAT and SemEval datasets the set of candidates is limited to a closed pre-defined set.
Relying on direction alone, while ignoring spatial distance is problematic when considering the entire vocabulary as candidates
because, we are likely to find candidates $d$ that have the same relation to $c$ as reflected by $\vec{a} - \vec{b}$.
For example, given the analogy \textit{man}:\textit{woman}::\textit{king}:\textit{?}, we are likely to recover feminine entities,
but not necessarily royal ones using PairDiff. On the other hand, in both SemEval and SAT datasets, the set of candidate answers already contains
the related candidates, leaving mainly the direction to be decided. For the remainder of the experiments described in the paper,
we use CosMult for evaluations on the Google dataset, whereas PairDiff is used for the SAT and SemEval datasets.
Results reported in Table~\ref{tbl:baselines} reveal that according to the binomial exact test with $95\%$ confidence
the proposed method statistically significantly outperforms GloVe, the current state-of-the-art word representation learning
method, on all three benchmark datasets.

To study the effect of the train dataset size on the performance of the proposed method,
following the procedure described in Section~\ref{sec:train:data},
we sample two balanced datasets containing respectively $10,000$ and $100,000$ instances.
 Figure~\ref{fig:datasetsize} shows the performance 
reported by the proposed method on the Google dataset.
We see that the overall performance increases with the dataset size,
and the gain is more for syntactic analogies. This result can be explained
considering that semantic relations are more rare compared to syntactic relations in the ukWaC corpus, a generic web crawl,
used in our experiments. However, the proposed train data selection method provides us with a potentially unlimited
source of positive and negative training instances which we can use to further improve the performance.

\begin{figure}[t]
\begin{center}
\includegraphics[width=90mm]{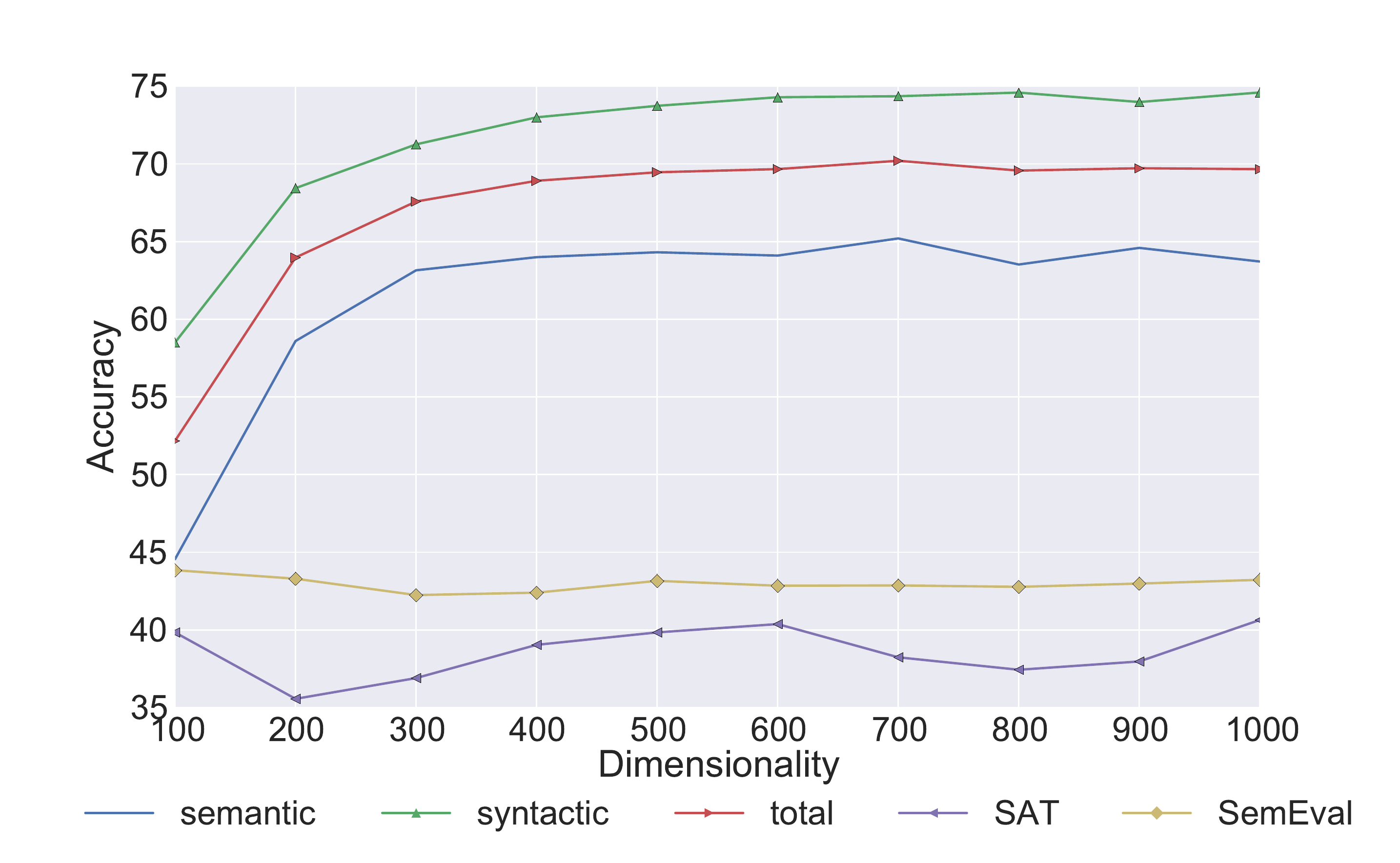}
\caption{Accuracy of the proposed method on benchmark datasets for dimensionalities of the word representations.}
\label{fig:dims}
\end{center}
\end{figure}

To study the effect of the dimensionality $d$ of the representation on the performance of the proposed method,
we hold the train data size fixed and produce word representations for different dimensionalities.
As shown in Figure~\ref{fig:dims}, the performance increases until around $600$ dimensions
on the Google, and the SAT datasets after which it stabilizes.
The performance on the SemEval dataset remains relatively unaffected by the dimensionality of the representation.

\section{Conclusions}

We proposed a method to learn word representations that embeds information related to semantic relations 
between words. A two step algorithm that alternates between pattern and  word representations
was proposed. The proposed method significantly outperforms the current
state-of-the-art word representation learning methods on three datasets containing proportional analogies.

Semantic relations that can be encoded as attributes in words are only a fraction of all types of semantic relations.
Whether we can accurately embed semantic relations that involve multiple entities, or
semantic relations that are only extrinsically and implicitly represented remains unknown.
We plan to explore these possibilities in our future work.

\bibliographystyle{named}
\bibliography{RelRep}

\end{document}